\documentclass{llncs}
\usepackage{times}
\usepackage{float}
\usepackage{footnote}
\usepackage{amsmath}
\usepackage{pgfplots}
\usepackage{indentfirst}
\usepackage{pbox}
\usepackage{multirow}
\usepackage{listings}
\usepackage{tabularx}
\usepackage{url}
\usepackage[boxruled]{algorithm2e}

\newcommand*\rfrac[2]{{}^{#1}\!/_{#2}}

\begin{document}
\title{Efficient Dodgson-Score Calculation Using Heuristics and Parallel Computing}
\author{Arne Recknagel\inst{1} \and Tarek R. Besold\inst{2}}
\institute{IBM Research and Development, \email{arecknag@gmail.com} \and The KRDB Research Centre, Free University of Bozen-Bolzano, \email{tarekrichard.besold@unibz.it}}
\maketitle
\begin{abstract}
Conflict of interest is the permanent companion of any population of agents (computational or biological). For that reason, the ability to compromise is of paramount importance, making voting a key element of societal mechanisms. A voting procedure often discussed in the literature and, due to its intuitiveness, also conceptually quite appealing is Charles Dodgson's scoring rule, basically using the respective closeness to being a Condorcet winner for evaluating competing alternatives. In this paper, we offer insights into the practical limits of algorithms computing the exact Dodgson scores from a number of votes. While the problem itself is theoretically intractable, this work proposes and analyses five different solutions which try distinct approaches to practically solve the issue in an effective manner. Additionally, three of the discussed procedures can be run in parallel which has the potential of drastically improving computational performance on the problem.
\end{abstract}

\section{Introduction: COMSOC and Dodgson's rule}
\label{chap:bat}
\label{sec:intro}
Voting is a common method for a group of individual agents, possibly having distinct goals, interests, and states of information, to coordinate and make a decision or express an opinion. The goal of voting is to reach a conclusion which leaves most of the voters content. In essence this is the only constraint any vote-interpretation procedure---from now on called a `rule'---needs to fulfil, and often serves as a means of evaluation whenever two rules disagree in their results. Over the last years, also researchers in Artificial Intelligence (AI) have taken growing interest in voting rules and choice processes in general, giving rise to the now flourishing AI subfield of Computational Social Choice (COMSOC) at the interface of social choice theory and computer science (cf., e.g., \cite{rothe_2015,brandt_2016}). COMSOC applies techniques developed in computer science to the study of social choice mechanisms (such as voting procedures or fair division algorithms), while also importing concepts from social choice theory into computing (such as social welfare orderings for multi-agent systems or network design).

A conceptually central voting rule within COMSOC is named after the mathematician Condorcet: When voting on different alternatives, a voter $v_i$ is asked to rank the list of alternatives according to preference (resulting in an order of preference $op_{v_i}$). 
If all orders of preference from all voters have been collected in a preference profile $pp$, the alternative which wins all pairwise comparisons (i.e., all hypothetical one-on-one contests with all other alternatives, based on which alternative is the one in the respective pair that the majority of voters rank higher in their orders of preference) is declared the winner of the election. Thus, if a winner is found she is preferred over any other alternative and is called a `(strong) Condorcet winner'. Unfortunately, a Condorcet winner does not always exist. Therefore, the notion of `Con\-dor\-cet-consistency' was introduced: The rule in question will select the Condorcet winner if one exists, but may otherwise also offer an alternative. 

\cite{dodgson_1876} refined Condorcet's rule: The number of swaps between adjacent alternatives is used as distance function between preference profiles, and the `Dodgson score' $sc_D$ of a candidate $x$ is the minimum distance from a profile in which $x$ is a Condorcet winner. The rule then selects an alternative globally minimising $sc_D$. This is Condorcet-consistent: If $x$ already is the Condorcet winner, $sc_D$ is zero and $x$ wins the election. Otherwise we give a ranking, indicating how far an alternative is from being a Condorcet winner.

However, a significant weakness of Dodgson's rule is its computational complexity, as the number of possible swaps grows exponentially with the number of voters and alternatives. 
Determining a winner under the Dodgson rule is complete for parallel access to $NP$ (\cite{hemaspaandra_1997}) and $W[1]$ hard parameterising by the number of voters (\cite{fellows_2010}).
Still, many problems are in theory known to be intractable but can nonetheless in practice be fairly efficiently solved. Therefore, this article empirically explores the practical capacities of problems which can be solved with a Dodgson rule implementation. 
The potential payoff is high: besides the direct application as voting procedure, an implementation 
constitutes a first step towards a basis for tests and comparisons of approximation algorithms like the ones proposed by \cite{mccabe_2008} or \cite{caragiannis_2014}. 
Therefore, we analyse several search heuristics and benchmark their performance---some 
include traditional approaches to improving search space requirements and/or speed, others are specifically tailored for this problem.\footnote{The source code of the project software is available from \url{https://sourceforge.net/projects/dodgsonscoring/}.}

\section{The Baseline Scorer}
\label{chap:baseline}
\label{sec:baselinescorer}

In order to reliably assess and compare the quality of different solutions to solving a Dodgson voting scheme, a baseline is required. In our baseline---as in all other approaches reported after it---the problem of finding the Dodgson score is treated as a search problem over the space of all possible profiles. The winning condition for a specific profile is having the alternative under consideration as Condorcet winner. Additionally, no other profile in the search space which also offers a Condorcet winner may have a lower Dodgson score.

Thus, the corresponding scorer needs to generate the search space, and then search through it until it can reliably satisfy the conditions for at least one solution. A first constraint can be implemented increasing computational efficiency. The search space will largely consist of solutions which are permutations with changes irrelevant or detrimental to the alternative in question. Since these cannot possibly be a solution, if recognised such preference profiles can directly be filtered out during the search space construction. The search space is generated and stored together with the Dodgson scores for each profile and a second algorithm checks for each of the profiles if it is a Condorcet winner. Once the whole search space has been evaluated, the solution with the minimal score becomes the final solution. 

The size of the search space and the speed of the Condorcet winner finding procedure are the two parameters relevant for assessing the algorithm's size requirements and speed. However, if it is evaluated in relation to the heuristics-enriched improvements presented later, we only need to look at the search space size, as the Condorcet winner finding algorithm is identical for all approaches. Searching for all solutions, with $n$ voters and $m$ alternatives, the size of the total space is $\Phi_\text{basic}(n,m) = \sum_{i=1}^{m} i^n$ in the worst and $\Phi_\text{basic}(n,m) \leq m!^{(\frac{n}{m}+1)} m$ in the base case (cf. \cite{recknagel_PICS}).

\section{The Depth-First Approach}
\label{chap:dfs}
\label{sec:dfsapproach}

The search space is treated as a tree. Even for the baseline it is generated with a recursive function, which already mimics a depth-first search (DFS) behaviour. The expected improvement resides in needed disc space, as in a DFS every node is handled as a new solution and the winning conditions can be tested locally. Thus, while the baseline needs memory in the range of $O(n^m)$, requirements for a DFS contain only the tree depth, i.e. $O(n)$. The overall size of the search space remains the same and, in a DFS, the entire search space has to be traversed in order to assure that a found solution indeed is optimal. Thus, also the best/worst-case scenarios are identical to the baseline.

From a practical perspective, the implemented DFS solution follows the general information theory approach of DFS very closely and manages to solve the problem without too specific alterations to the core of the algorithm. It can be argued that DFS forms an upper bound of sorts on the `real' baseline for comparing Dodgson scores in this regard. Therefore, later implementations will also be benchmarked against it.

\section{The Uniform-Cost Approach}
\label{chap:ucs}
\label{sec:ucsapproach}

In finding the Dodgson score, obtaining a solution by itself is not sufficient unless it also is a minimal one. A classical DFS approach puts no useful preference over which path to chose and when to backtrack, as it only keeps track of local data. Therefore, a DFS needs to traverse the entire search space as any found solution is useless unless all others have also been seen. In contrast, uniform-cost search (UCS) is based on a breadth-first search (BFS). UCS can be performed if the edges of the tree are weighted. 
Keeping track of the scores (which can be designed to act as costs) and exploring the search space while keeping them minimal comes naturally to it. By exploring the nodes first which have the lowest cost, we can assert that each currently examined node is a global minimum for the whole search space.

Still, there are drawbacks. Keeping global data (as required for BFS-type mechanisms) may help speed up the algorithm. This is useful as the problems get bigger, but bigger problems also produce more data. Another problem is that preference profiles, when thought of in terms of change over time, are inefficient to work with. That is why `swap profiles' are used here and in the following two implementations. The original profile is only stored once, together with the alternative under examination. All new profiles can be thought of as variants of that profile in which only the position of the current alternative per agent is changed. We obtain a vector of integers of cardinality $n$ (i.e., number of agents) per node instead of the full profile. The sum of all values in the vector is the Dodgson score, simplifying comparisons between nodes and facilitating sorting.

The speed of this algorithm depends solely on the moment the first solution appears in the tree. General best and worst case profiles and the resulting search space size are still the same as in the baseline scenario. Still, since the procedure stops at the first possible solution in practice the latter's position becomes decisive. The search space is traversed in an ordered fashion and can be thought of as sorted according to Dodgson scores. Using UCS, only the part of $\Phi$ up to the first solution is checked, giving the actually traversed search space $\{C \subseteq \Phi \mid  sc_D(c) \le sc_D(\phi),\forall c \in C, \forall\phi  \in \Phi \backslash C\}$. 

\emph{Actual worst case:}
A worst-case scenario for an alternative $x$ arises if all $n$ voters submit identical preference rankings listing $x$ last. For $x$ to become the Condorcet winner---since in direct comparison the approval for each alternative has to be worse than for $x$---it would need to be at the top for more than half of the voting agents. Thus, when procedurally traversing the set of agents in an ordered way (e.g., sorting agents by increasing index) and denoting $x$'s position in the orders of preference with $i_x$, for a particular $x$ the number of swap permutations/solutions to be checked is upper bounded by $C(x) = (i_x)^{\lceil\frac{n}{2}\rceil}$, and by $C({all}) = \sum_{i=1}^{m} (i)^{\lceil\frac{n}{2}\rceil}$ for all alternatives.
With the number of alternatives $m$ ranging from 1 to 10 and the number of agents $n=5$, experiments give the values in Table~\ref{table:11} for the total search space $\Phi(n, m)$, the actually traversed space $C(n, m)$, 
and the ratio $\frac{C}{\Phi}$ in percent. 

\begin{table}
\centering
\footnotesize
\begin{tabular}{r r r r r}
	\hline
	$n,m$ & $\Phi(n,m)$ & $C(n,m)$ & $C/\Phi$ \\ \hline
	1, 5 & 1 & 1 & 100.0\% \\
	2, 5 & 33 & 9 & 27.3\% \\
	3, 5 & 276 & 36 & 13.0\% \\
	4, 5 & 1300 & 100 & 7.7\% \\
	5, 5 & 4425 & 225 & 5.1\% \\
	6, 5 & 12201 & 441 & 3.6\% \\
	7, 5 & 29008 & 784 & 2.7\% \\
	8, 5 & 61776 & 1296 & 2.1\% \\
	9, 5 & 120825 & 2025 & 1.7\% \\
	10, 5 & 220825 & 3025 & 1.4\% \\ \hline

 \end{tabular}
 \caption{Exploration rate of the overall search space in the UCS worst-case setting.}
\label{table:11}
\end{table}

In a randomised setup, the worst case seems rather unlikely to occur. 
However, `worst-case scenarios' might not be as rare in a real-world setup due to a bias among alternatives: An alternative which is (dis)liked by one is often also (dis)liked by others. Sect. \ref{chap:threading} includes a proposition on how to address this tendency of realistic problems to shift towards the worst-case scenario.

\emph{Actual best case:}
We again estimate $C$ in relation to $\Phi$. Fulfilling the Condorcet condition is more complex for the best case as the agents' preferences are not interchangeable. As the search space is traversed in an ordered way, if the first winning profile has a Dodgson score of $x$, after that at most all profiles with a score $\leq x$ have to be considered. The minimal Dodgson score for a best-case scenario can in general be lower bounded as $min(sc_{D}(x)) \geq (\frac{n}{m}-1)\sum_{i=1}^{\rfrac{n}{2}}i$.  The ratio $\frac{C}{\Phi}$ in Table~\ref{table:12} shows how much of the space maximally has to be explored using the pruning criterion in a $n=m$ setting.

\begin{table}
\footnotesize
\centering
\begin{tabular}{l r r r}
	\hline
	$n = m$ & $\Phi(n,m)$ & $C(n,m)$ & $C/\Phi$ \\ \hline
	1 & 1 & 1 & 100.0\% \\
	2 & 2 & 2 & 100.0\% \\
	3 & 6 & 3 & 50.0\% \\
	4 & 24 & 15 & 62.5\% \\
	5 & 120 & 29 & 24.2\% \\
	6 & 720 & 259 & 36.0\% \\
	7 & 5040 & 602 & 11.9\% \\
	8 & 40320 & 8039 & 19.9\% \\
	9 & 362880 & 21671 & 6.0\% \\
	10 & 3628800 & 392588 & 10.8\% \\
	11 & 39916600 & 1200900 & 3.0\% \\
	12 & 479001600 & 27770328 & 5.8\% \\ \hline
 \end{tabular}
\caption{Comparison between overall and actual search space in the UCS best-case setting (with $n=m$).}
\label{table:12}
\end{table}

\section{Smart Caching and Iterative Cost Raise}
\label{chap:scicr}
\label{sec:scicrapproach}

Smart Caching (SC) and Iterative Cost Raise (ICR) are alternatives for semi-informed search with simultaneous pruning of the search space. In the following, the `swap space' is a representation of the search space defined by the initial preference profile with the currently examined alternative, the swap profiles representing the possible solutions, and a position table mapping swap profiles back into a preference profile. Using the swap space, we can more easily sort potential solutions as it is possible to targetedly generate swap profiles with a specific Dodgson score by distributing a summand partition of the score among the $n$ agents. 

For both algorithms we start with the lowest possible Dodgson score, and generate all possible swap profiles from it. If no solution is found, we proceed with the next higher Dodgson score, etc. Thus the name for SC: We are caching a certain subspace of all possibilities in order to search a valid solution among them. ICR, described in pseudo code in Algorithm \ref{ICR_pseudo_code}, is an improvement to SC in the way that DFS is to the baseline.

\begin{algorithm}
\SetKwFunction{Search}{ICRSearch}
\SetKwFunction{CW}{CondorcetWinner}
\SetKwInput{Create}{create}
\Begin(Search Code){
  \Search{$pp = v_{1 1} v_{1 2} \ldots v_{i j} \ldots v_{n m}, a \in \lbrace a_1 a_2 \ldots a_m \rbrace$}\\
  \KwData{A preference profile and one of the alternatives present in it}
  \KwResult{The $Dodgson Score$ of the alternative in question}
  \BlankLine
  sc$_D\leftarrow$ 0\;
  \While{True}{
   \Create{Permutor(sc$_D$, pp, a);}
   \While{Permutor.hasNext()}{
    next $\leftarrow$ Permutor.next() $\times$ pp\;
    \If{\CW{next, a}}{
     \KwRet{sc$_D$;}
    }
   }
  sc$_D\leftarrow$ sc$_D+1$\;
  }
 }
 \Begin(Permutator description){Iterates over all possible upward swaps given a maximum number of swaps (i.e. the current Dodgson score), a preference profile, and the alternative in question. The resulting list of integers will, when applied to the preference profile, produce its corresponding permutation.
 }
\caption{The ICR algorithm.\label{ICR_pseudo_code}}
\end{algorithm}

While especially ICR appears similar to iterative deepening DFS, SC's and ICR's core attribute is fundamentally different in that we can choose precisely which states to explore, meaning no previous work needs to be repeated. ICR's immediate evaluation has another advantage. Keeping track of the global minimum means that the the first encountered valid solution stops the summand partition algorithm, leading to an additional gain in speed. Finally, the position table is critical to both approaches as it makes sure that no impossible profile is generated.
As currently a recursive generation algorithm is used, the gain from just one early stop can be immense, avoiding all consecutive faulty builds.

Abstractly, the search space is conceptualised as a layered set with no  connectivity between its elements instead of a tree. The size of the ultimately generated space depends on when a solution is found. The number of profiles measured against their scores in estimation behaves like a quadratic function, causing finding an early solution to yield significant payoff. Partly similar to the baseline, caching the search space and searching for a solution is separated into different algorithms. Still, even in a bad case only a fraction of the total space is cached at any point in time, and the swap profiles need less memory for storing.

While the search could be stopped with the first winner, the current implementation finishes only after checking the complete corresponding layer in order to collect all minimal solutions. This facilitates the comparison of solutions with the other scorers. Additionally, in some settings it could also be important to find all solutions, and the computational resource requirements stay within the same equivalence class: $O(n\cdot(1-p))$, with layer size $n$ and $p=[0..1]$ the probability of finding a solution at any point in the layer. $O(n)$ is the amount of work required for searching the complete layer.

\section{Multi-processing/-threading}
\label{chap:threading}
\label{sec:multi}

Much of the complexity of the naive Dodgson rule results from the need to resolve the tournament instead of just having to obtain the score for one alternative---even if an individual score has been found, the most important information is whether the tournament has been won with that score. However, for UCS, SC, and ICR we can assert the minimality of a solution. This enables (at least) three alternative strategies:

\noindent 1.) Run all alternative searches simultaneously. The core algorithms take the alternative as parameter and are called $m$ times. 
Once a solution is found, all algorithms with a greater current Dodgson score can halt.\\
\noindent 2.) Run the solutions procedurally, but stop as soon as an already computed Dodgson score is exceeded. As further improvement estimate which solutions will produce a low score and run those first. An easy heuristic is to sum up their distances to the top for all voters, i.e. sort them using a Borda count (\cite{ratliff_2002}).\footnote{In a Borda count voters rank alternatives in order of preference. The outcome is determined by assigning each alternative, for each ballot, a number of points corresponding to the number of candidates ranked lower. The alternative with the highest number of points, summed up over all cast votes, is the winner.}\\
\noindent 3.) Do both.

Consider a $k \times k$ worst-case scenario with all preference orderings being $(a_1<a_2<\cdots<a_{k-1}<a_k)$ (i.e., $a_1$ being the most and $a_k$ the least preferred alternative).
Applying the heuristic, this becomes a best case: BordaCount$(a_1)$ returns the highest value of all alternatives, and running $a_1$ first produces an instant solution and halts the computation. The balanced former best case turns into a worst case in which all solutions need to be fully computed as none is better than the next. The former best case is now the (comparatively manageable) worst case; at the cost of getting only the Dodgson winner performance significantly improved.

A best case occurs each time the profile contains a Condorcet winner, as this leads to an immediate halt. The more agents and alternatives there are, the lower the likelihood that this occurs (cf. \cite{gehrlein_1999}). The worst case, with each alternative being perfectly symmetric to all others, is in comparison a lot less probable. Only one alternative breaking the pattern results in a much more favourable case.

In summary, the total speed solely depends on the easiest to compute alternative. Since alternatives do not exist independent of each other, there cannot be only bad ones; at worst we can have many mediocre ones. If no solution is found, we halt after a set time and assert that no alternative is better than the latest score.

\section{Benchmarking \& Results}
\label{chap:benchmarking}

The preference profiles used for benchmarking are generated pseudo-randomly. When benchmarking several scorers, the initialisation seed for the pseudo-random procedure is saved and re-used to guarantee comparability of the results.
Impartial culture for all agents is assumed in benchmarking (i.e., one agent's voting behaviour gives no information about other agents' preferences). While this makes most sense for testing, in a more realistic setup agents will more often than not have similar (or at least correlated) preferences. For the standard versions of all algorithms this approaches the worst, for the threaded variants the best case.

\emph{Average Performance and Maximum Range:} The scorers were first tested on average performance. The chosen problem size constitutes a tradeoff between minimising the influence of rounding errors or performance lows and maximising the amount of collected data within limited time. The unit of measure is clock ticks, with one tick taking one millisecond on the used system, and the results of 1000 runs using the standard algorithms on a $8\times5$ preference profile are given in Table~\ref{table:16} (best performance is bolded, $\sigma$ indicates the standard deviation). The multiprocessing variants of UCS, SC, and ICR were also tested on the same problem. An $8\times5$ profile with 1000 runs was chosen, and baseline and DFS were run as before. The results are given in Table~\ref{table:17}.

\begin{table}
\centering
\footnotesize
\begin{tabular}{l | c c c | c c r}
	\hline
  $scorer$ & $min$ & $median$ & $max$  & $mean$ & $\sigma$ & $avg. calls$ \\ \hline
Base   & 6377  &  15805   & 137865 & 18135  & 10759    & \textbf{16425}  \\
DFS        &  847  &   1608   &  \textbf{13564} &  2028  &  \textbf{1193}    & 16527  \\
UCS        &  204  &   1685   &  85254 &  3693  &  6241    & 17540  \\
SC	   	   &   76  &   1642   & 320828 &  6313  & 20020    & 76409  \\
ICR      & \textbf{50}  &    \textbf{552}   &  27557 &  \textbf{1202}  &  2042    & 55706  \\ \hline
\end{tabular}
\caption{Average performance of standard algorithms.
}
\label{table:16}
\end{table}

\begin{table}
\centering
\footnotesize
\begin{tabular}{l | c c c | c c r}
	\hline
  $scorer$ & $min$ & $median$ & $max$  & $mean$ & $\sigma$ & $avg. calls$ \\ \hline
Base   & 7106  &  15090   & 110112 & 18016  &  9707    & 16536       \\
DFS        &  913  &   1836   &  10712 &  2167  &  1169    & 16558  \\
UCS        &    3  &     14   &    145 &    20  &    21    &   \textbf{225}  \\
SC	   	   &    \textbf{0}  &      3   &     \textbf{55} &     \textbf{5}  &     \textbf{6}    &   434  \\
ICR      &    \textbf{0}  &      \textbf{2}   &     67 &     \textbf{5}  &     \textbf{6}    &   407  \\ \hline
\end{tabular}
\caption{Average performance of threaded algorithms.}
\label{table:17}
\end{table}

\begin{figure*}
\centering
\begin{minipage}{.4\textwidth}
  \centering
  \includegraphics[width = \linewidth]{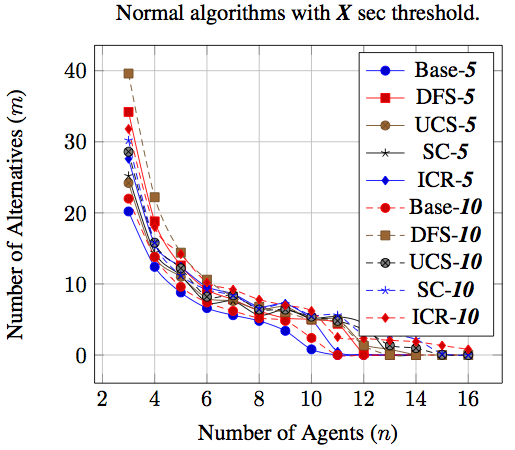}
\end{minipage}
\begin{minipage}{.4\textwidth}
  \centering
  \includegraphics[width = \linewidth]{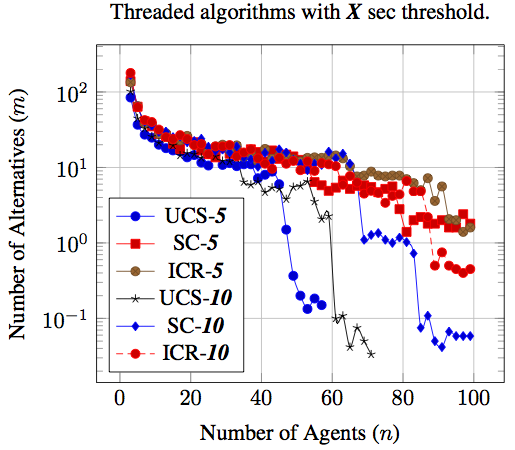}
\end{minipage}
\caption{Experimental range comparison between standard and threaded algorithms on Dodgson's rule.}
\label{fig1}
\end{figure*}

In order to assess the impact of exponential growth, each algorithm was run with increasingly complex problems. $n$ is constant and $m$ is incremented each time a solution is found. When exceeding a preset time window, the final value of $m$ is stored and the process is repeated with $n+1$. The process is stopped at the first $n$ with $m \leq 4$. The averaged results over five runs (for two different time thresholds) are given in Fig.~\ref{fig1}.\footnote{Similar to \cite{gehrlein_1999}, even values for $n$ were omitted as the performance of multiprocessing algorithms is influenced by the probability of there being a Condorcet winner. Even numbers of agents make this significantly less likely due to ties, resulting in the graph spiking after every other value.}

\emph{Sequential Algorithm Results:} While theoretically expected as significant, the data in Table \ref{table:16} shows only minor gains for the sophisticated algorithms. A speed\-up by the factor $3-17$ appears negligible, given the marginal gain in size of solvable problems. In an exemplary maximum range testing, the baseline can only solve problems up to sizes $3\times22$ to $9\times5$ reliably. UCS and SC do not beat a simple DFS in the average performance test. ICR does, but by far less than theoretically suggested. All three advanced algorithms have significantly higher max-values than DFS with bad cases occurring often enough to cripple the overall outcome.

Several reasons for the low per\-for\-mance are imaginable: While UCS and SC are as fast as or faster than DFS in most cases, their average performance is dragged down by very time-consuming bad cases, as reflected by the high $\sigma$. Also, best cases are rare. If a single alternative from the whole profile leans towards a higher Dodgson score, the whole process slows down since the amount of computational resources required grows exponentially with the highest score of the preference profile. Good cases, however, are solved very quickly as indicated by the low minima. Third, in UCS, even if the amount of explored leaves is considerably low, the hidden overhead are all the paths which lead to them. In any tree with depth $n$ and branching factor $m$ the number of total nodes is $\sum_{i=1}^n m^i$, so $m^n$ leafs can have $\sum_{i=1}^{n-1} m^i$ hidden nodes. These were not present in DFS: only changes in the actual profile were explored. The overhead for SC and ICR might be even bigger, as after each layer the intermediate swap profiles need to be re-generated. Finally, using swap profiles possibly impairs performance more than assumed. As DFS directly handles preference profiles as node data it can skip the transformation for the Condorcet checker.

Comparing loop iteration counts, UCS needs on average about $1.6$ times longer per checked node than DFS, SC on the other hand only $0.7$ times, and ICR even only $0.16$ times. The number of core function calls vastly differs. UCS has little more than the baseline and DFS, but as SC and ICR use intermediate profiles the iteration count is considerably higher. 
For building one permutation with $n$ agents, they need a total of $n-1$ intermediate profiles. Since many intermediate profiles are shared among similar profiles, the final number of checked instances is lower than $n-1$ times the expected number, but still can be quite high.

\emph{Multiprocessing Algorithm Results:} Weakening the winning condition to only search the Dodgson winner, and having multiple threads search simultaneously, lowers the worst case to the former best case. Due to the drastic reduction of loop iterations for finding a solution, corresponding scorers are $1000-4000$ times faster than the baseline in the average performance tests (this difference further grows with the problem size). The maximum range of solvable problems reflects this. In our setup the best multithreading approach solves preference profiles up to sizes of $3\times172$ to $92\times4$. 

\section{Related and Future Work \& Conclusion}
\label{chap:conclusion}

A different approach to solving Dodgson's rule has been described by \cite{homan_2009}. They show that for a sufficiently high number of voters and only limitedly many alternatives a fairly simple, polynomial-time greedy algorithm very frequently finds the Dodgson winners. If a small degree of uncertainty is tolerated, provided with a Dodgson election and one of the alternatives, their {\tt GreedyWinner} algorithm outputs whether or not it considers the alternative a Dodgson winner and one of the two confidence values ``definitely'' or ``maybe''. 
If ``definitely'' is returned as second component, the part of the answer concerning the status of the alternative as Dodgson winner (or not) is provably correct. This approach differs from the ones taken above in that \cite{homan_2009}'s algorithm is greedily heuristic with the correctness probability increasing with the ratio of voters to alternatives, contrasting with our precise mechanisms considered above. An implementation and experimental comparison between our alternatives and the {\tt GreedyWinner} is future work---as is encoding the problem as a mixed-integer program and running it on a corresponding solver.

Summarising, our initial goal was to work towards algorithms solving non-trivial instances of Dodgson's rule. Besides being applicable by themselves, if the handled problem size was big enough, the algorithms could 
help in 
developing rules approximating the Dodgson score. Both goals were met. Especially if only the Dodgson winner is sought, the threaded implementations can solve a wide range of problem sizes in reasonable time. Also, we managed to effectively reduce the problem size for the constrained case. Still, the time needed for finding a Dodgson winner might nonetheless be too high. However, approximations to the Dodgson score are justifiable, and since computation can be stopped at any point, we can extract the maximal score which was tested and conclude that no solution with a lower score exists. This is especially interesting for multi-agent systems, which can then decide to resort to other means of decision-making, since casting a vote was inconclusive.


\bibliographystyle{splncs}
\bibliography{dodgson_kijournal_2016}

\begin{thebibliography}{10}

\bibitem{rothe_2015}
Rothe, J., ed.:
\newblock {Economics and Computation. An Introduction to Algorithmic Game
  Theory, Computational Social Choice, and Fair Division}.
\newblock Springer Texts in Business and Economics. Springer (2015)

\bibitem{brandt_2016}
Brandt, F., Conitzer, V., Endriss, U., Lang, J., Procaccia, A., eds.:
\newblock {Handbook of Computational Social Choice}.
\newblock Cambridge University Press (2016)

\bibitem{dodgson_1876}
Dodgson, C.:
\newblock Reprint of ``{A} method of taking votes on more than two issues''.
\newblock In McLean, I., Urken, A., eds.: Classics of social choice, The
  University of Michigan Press (1876) Reprinted in 1995.

\bibitem{hemaspaandra_1997}
Hemaspaandra, E., Hemaspaandra, L.A., Rothe, J.:
\newblock {Exact Analysis of Dodgson Elections: Lewis Carroll's 1876 Voting
  System is Complete for Parallel Access to NP}.
\newblock Journal of the Association for Computing Machinery \textbf{44} (1997)
   806--825

\bibitem{fellows_2010}
Fellows, M., Jansen, B.M.P., Lokshtanov, D., Rosamond, F.A., Saurabh, S.:
\newblock {Determining the Winner of a Dodgson Election is Hard}.
\newblock In: IARCS Annual Conference on Foundations of Software Technology and
  Theoretical Computer Science. Volume~8 of Leibniz International Proceedings
  in Informatics (LIPIcs)., Schloss Dagstuhl--LZI (2010)  459--468

\bibitem{mccabe_2008}
McCabe-Dansted, J., Pritchard, G., Slinko, A.:
\newblock {Approximability of Dodgson's Rule}.
\newblock Social Choice and Welfare \textbf{31} (2008)  311--330

\bibitem{caragiannis_2014}
Caragiannis, I., Kaklamanis, C., Karanikolas, N., Procaccia, A.:
\newblock {Socially Desirable Approximations for Dodgson's Voting Rule}.
\newblock ACM Transactions on Algorithms \textbf{10} (2014)

\bibitem{recknagel_PICS}
Recknagel, A.:
\newblock An Approach to Efficiently Calculating Dodgson-Scores Using
  Heuristics and Parallel Computing. Volume 01-2015 of Publications of the
  Institute of Cognitive Science (PICS).
\newblock Institute of Cognitive Science, Osnabr\"uck (2015)

\bibitem{ratliff_2002}
Ratliff, T.C.:
\newblock {A comparison of Dodgson's method and the Borda count}.
\newblock Economic Theory \textbf{20}(2) (2002)  357--372

\bibitem{gehrlein_1999}
Gehrlein, W.V.:
\newblock {Approximating the Probability that a Condorcet Winner Exists}.
\newblock In: Proceedings of the National Decision Sciences Institute. (1999)
  626--628

\bibitem{homan_2009}
Homan, C.M., Hemaspaandra, L.A.:
\newblock {Guarantees for the success frequency of an algorithm for finding
  Dodgson-election winners}.
\newblock Journal of Heuristics \textbf{15}(4) (2009)  403--423

\end{thebibliography}

\end{document}